\documentclass[10pt,twocolumn,letterpaper]{article}

\usepackage[pagenumbers]{cvpr} 

\usepackage{graphicx}
\usepackage{amsmath}
\usepackage{amssymb}
\usepackage{booktabs}
\usepackage{siunitx}
\usepackage{cite}
\usepackage{ctable}
\usepackage{hhline}
\usepackage{booktabs} 
\usepackage{subcaption}
\usepackage{csquotes}
\usepackage{multirow}
\usepackage{pifont}
\usepackage[super]{nth}
\usepackage{wrapfig}
\usepackage{enumitem}

\definecolor{darkergreen}{RGB}{21, 152, 56}
\definecolor{red2}{RGB}{252, 54, 65}
\newcommand{\cmark}{\textcolor{darkergreen}{\ding{51}}}%
\newcommand{\xmark}{\textcolor{red2}{\ding{55}}}%

\newcolumntype{L}[1]{>{\raggedright\let\newline\\\arraybackslash\hspace{0pt}}m{#1}}
\newcolumntype{C}[1]{>{\centering\let\newline\\\arraybackslash\hspace{0pt}}m{#1}}
\newcolumntype{R}[1]{>{\raggedleft\let\newline\\\arraybackslash\hspace{0pt}}m{#1}}

\usepackage[pagebackref,breaklinks,colorlinks]{hyperref}

\usepackage[capitalize]{cleveref}
\crefname{section}{Sec.}{Secs.}
\Crefname{section}{Section}{Sections}
\Crefname{table}{Table}{Tables}
\crefname{table}{Tab.}{Tabs.}

\usepackage[accsupp]{axessibility}  

\def\confName{CVPR}
\def\confYear{2023}

\begin{document}

\title{Bringing Inputs to Shared Domains for \\ 3D Interacting Hands Recovery in the Wild}

\author{Gyeongsik Moon\\
Meta Reality Labs \\
{\tt\small mks0601@gmail.com}
}
\maketitle

\begin{abstract}
Despite recent achievements, existing 3D interacting hands recovery methods have shown results mainly on motion capture (MoCap) environments, not on in-the-wild (ITW) ones.
This is because collecting 3D interacting hands data in the wild is extremely challenging, even for the 2D data.
We present InterWild, which brings MoCap and ITW samples to shared domains for robust 3D interacting hands recovery in the wild with a limited amount of ITW 2D/3D interacting hands data.
3D interacting hands recovery consists of two sub-problems: 1) 3D recovery of each hand and 2) 3D relative translation recovery between two hands.
For the first sub-problem, we bring MoCap and ITW samples to a shared 2D scale space.
Although ITW datasets provide a limited amount of 2D/3D interacting hands, they contain large-scale 2D single hand data.
Motivated by this, we use a single hand image as an input for the first sub-problem regardless of whether two hands are interacting.
Hence, interacting hands of MoCap datasets are brought to the 2D scale space of single hands of ITW datasets.
For the second sub-problem, we bring MoCap and ITW samples to a shared appearance-invariant space.
Unlike the first sub-problem, 2D labels of ITW datasets are not helpful for the second sub-problem due to the 3D translation's ambiguity.
Hence, instead of relying on ITW samples, we amplify the generalizability of MoCap samples by taking only a geometric feature without an image as an input for the second sub-problem.
As the geometric feature is invariant to appearances, MoCap and ITW samples do not suffer from a huge appearance gap between the two datasets.
The code is publicly available\footnote{\url{https://github.com/facebookresearch/InterWild}}.
\end{abstract}

\begin{figure}[t]
\begin{center}
\includegraphics[width=\linewidth]{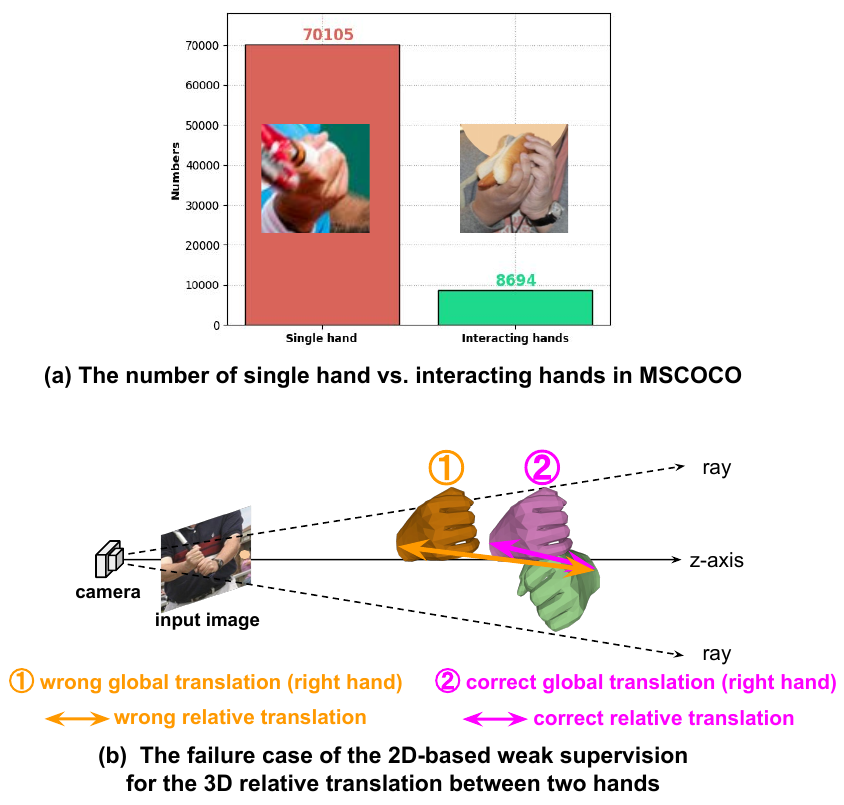}
\end{center}
\vspace*{-5mm}
\caption{
(a) Only a very small amount of 2D interacting hands are available in ITW datasets despite a relaxed threshold of intersection-over-union (IoU).
We consider two hands are interacting if the IoU between the two hands' boxes is bigger than 0.1.
(b) The 2D-based weak supervision from ITW datasets often results in a wrong 3D relative translation (\textcolor{orange}{\textbf{the orange arrow}}).
}
\vspace*{-3mm}
\label{fig:motivation}
\end{figure}

\section{Introduction}

3D interacting hands recovery aims to reconstruct a single person's interacting right and left hands in the 3D space.
The recent introduction of a large-scale motion capture (MoCap) dataset~\cite{moon2020interhand2} motivated many 3D interacting hands recovery methods~\cite{zhang2021interacting,rong2021monocular,li2022interacting,di2022lwa,hampali2022keypoint}.

Although they have shown robust results on MoCap datasets, none of them explicitly tackled robustness on in-the-wild (ITW) datasets.
Simply training networks on MoCap datasets and testing them on ITW datasets results in unstable results due to a huge domain gap between MoCap and ITW datasets.
The most representative domain gap is an appearance gap.
For example, images in InterHand2.6M~\cite{moon2020interhand2} (IH2.6M) have black backgrounds and artificial illuminations, far from those of ITW datasets.
The fundamental solution for this is collecting large-scale ITW data with 3D groundtruths (GTs); however, this is extremely challenging.
For example, capturing 3D data requires tens of calibrated and synchronized cameras.
Preparing such a setup at diverse places in the wild requires a huge amount of manual effort.
Furthermore, collecting even large-scale ITW 2D interacting hand data with manual annotation is greatly challenging due to the severe occlusions and self-similarities.
Due to such challenges, there is no large-scale ITW 2D/3D interacting hand dataset.

Nevertheless, ITW datasets provide large-scale 2D \emph{single-hand} data, as shown in Fig.~\ref{fig:motivation} (a).
Utilizing such large-scale 2D single-hand data of ITW datasets can be an orthogonal research direction to the 2D/3D interacting hands data collection in the wild.
Mixed-batch training is the most dominant approach to utilize 2D data of ITW datasets for the 3D human recovery~\cite{kolotouros2019learning,moon2020i2l,rong2021frankmocap,kocabas2021pare,moon2022hand4whole,choi2022learning}.
During the mixed-batch training, half of the samples in a mini-batch are taken from MoCap datasets and the rest of the samples from ITW datasets.
The MoCap samples are fully supervised with 3D GTs, and the ITW samples are weakly supervised with 2D GTs.
The ITW samples make networks exposed to diverse appearances, which leads to successful generalization to unseen ITW images.
The 2D-based weak supervision is enabled by the MANO~\cite{romero2017embodied} hand model, which produces a 3D hand mesh from pose and shape parameters in a differentiable way.
To be specific, 3D joint coordinates, extracted from the 3D mesh, are projected to the 2D space using an estimated 3D global translation (\textit{i.e.}, 3D translation from the camera to the hand root joint) and fixed virtual camera intrinsics.
Then, the projected 2D joint coordinates are supervised with the 2D GTs.
In this way, the 2D GTs weakly supervise MANO parameters, which can make all vertices of the 3D mesh fit to the 2D GTs.

\begin{figure}[t]
\begin{center}
\includegraphics[width=0.9\linewidth]{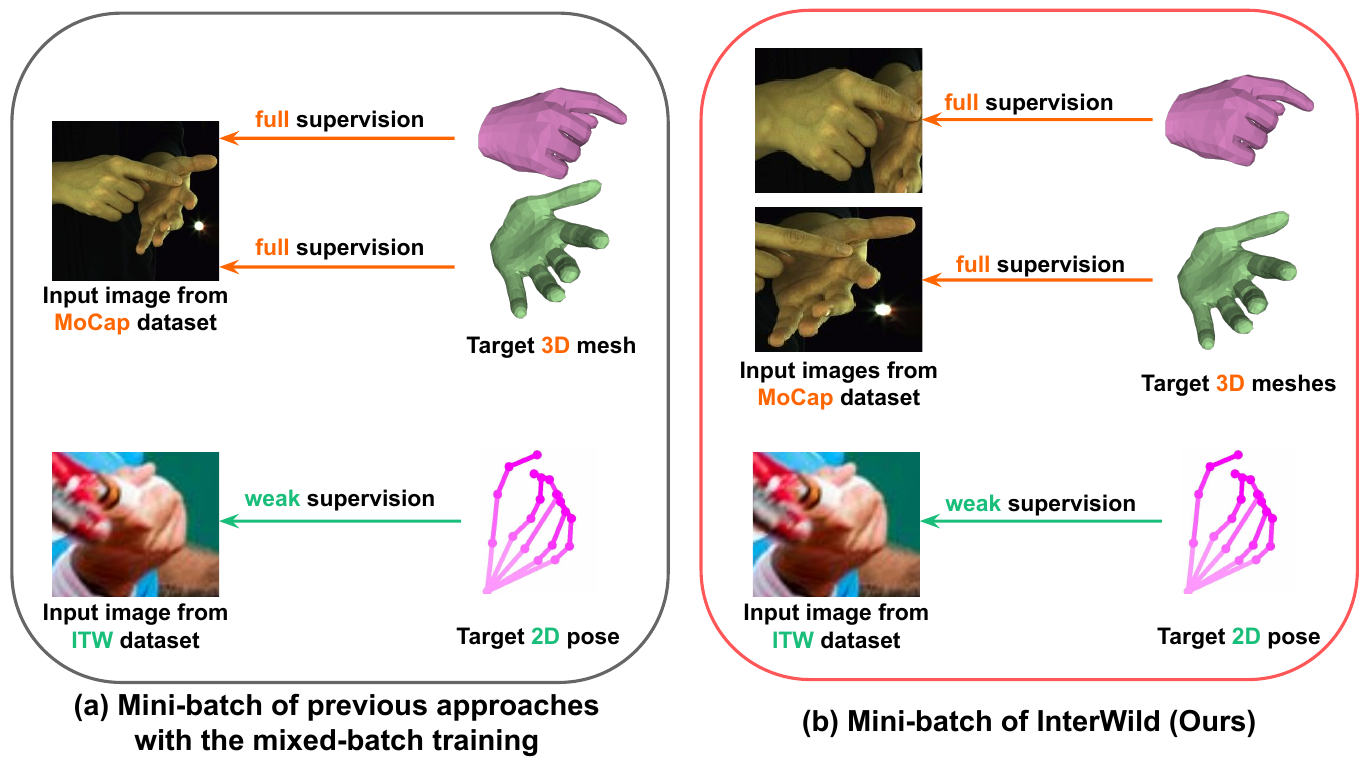}
\end{center}
\vspace*{-5mm}
\caption{
Mini-batch comparison for the first sub-problem (\textit{i.e.}, estimation of separate 3D meshes of left and right hands).
}
\vspace*{-3mm}
\label{fig:shnet_compare_prev}
\end{figure}

However, naively re-training networks of
 previous 3D interacting hands recovery methods~\cite{zhang2021interacting,rong2021monocular,li2022interacting,di2022lwa,hampali2022keypoint} with the mixed-batch training does not result in robust results.
3D interacting hands mesh recovery consists of two sub-problems: 1) estimation of separate 3D right and left hands and 2) estimation of 3D relative translation between two hands.
For the first sub-problem, previous works~\cite{zhang2021interacting,rong2021monocular,li2022interacting,di2022lwa,hampali2022keypoint} take an image of two hands when hands are interacting, and an image of a single hand when hands are not interacting (Fig.~\ref{fig:shnet_compare_prev} (a)).
As ITW datasets mostly contain single-hand data (Fig.~\ref{fig:motivation} (a)), most samples from ITW datasets contain a single hand during the mixed-batch training.
The problem is that the images of two hands from MoCap datasets have very different 2D hand scale distribution compared to that of single-hand images from ITW datasets, as shown in Fig.~\ref{fig:sh_ih_dist_compare}.
For example, when two hands are included in the input image, the 2D scale of each hand is much smaller than that from a cropped image of a single hand.

\begin{figure}[t]
\begin{center}
\includegraphics[width=\linewidth]{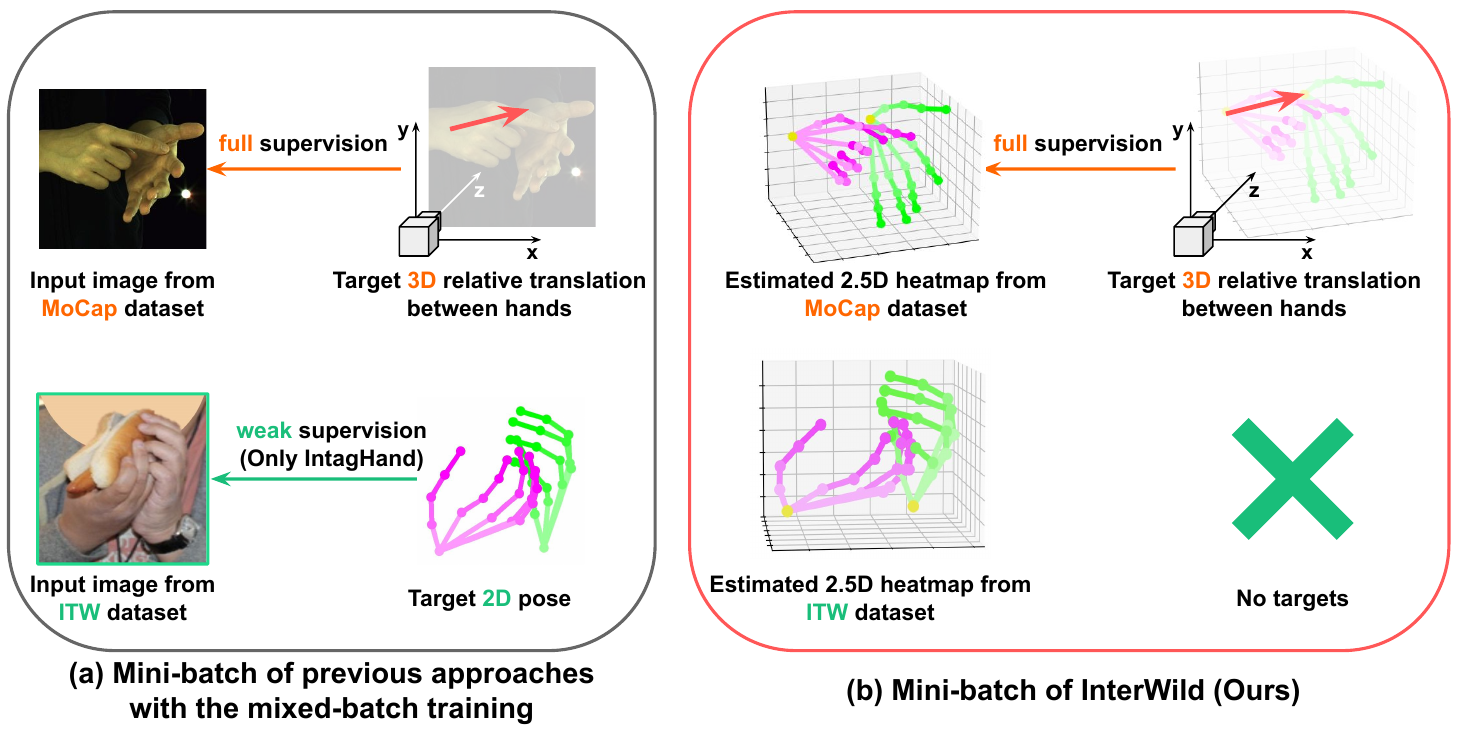}
\end{center}
\vspace*{-5mm}
\caption{
Mini-batch comparison for the second sub-problem (\textit{i.e.}, estimation of 3D relative translation between two hands).
}
\vspace*{-3mm}
\label{fig:transnet_compare_prev}
\end{figure}

Unlike the first sub-problem, the second sub-problem hardly gets benefits from the 2D-based weak supervision from ITW datasets.
Fig.~\ref{fig:motivation} (b) shows the failure case of the 2D-based weak supervision.
When the 3D global translation, estimated for the 2D-based weak supervision, is wrong (\textcolor{orange}{\textbf{\textcircled{\raisebox{-0.9pt}{1}}}} in the figure), the 3D relative translation is supervised to be wrong one (\textcolor{orange}{\textbf{the orange arrow}} in the figure).
The wrong 3D global translation also can happen to the first sub-problem; however, the critical difference is that the 3D scale of the 3D relative translation (\textit{i.e.}, output of the second sub-problem) is very weakly constrained, while the 3D scale of hands (\textit{i.e.}, output of the first sub-problem) are strongly constrained by the shape parameter of MANO~\cite{romero2017embodied}.
For example, we can place two hands close or far freely based on how they are interacting with each other, while the size of adults' hands is usually around 15 cm.
As there is no such strong constraint to the relative translation, the relative translation can be an arbitrary value and is prone to wrong supervision (\textcolor{orange}{\textbf{the orange arrow}} in the figure).
Please note that estimating a 3D global translation from a single image involves a high ambiguity and can be often wrong as the camera position is not provided in the input image.
In this regard, we observed that the 2D-based weak supervision for the 3D relative translation, used in IntagHand~\cite{li2022interacting} (Fig.~\ref{fig:transnet_compare_prev} (a)) deteriorates results, which is shown in the experimental section.
However, without the 2D-based weak supervision of ITW datasets, the network is trained only on images of MoCap datasets like previous works~\cite{zhang2021interacting,rong2021monocular,di2022lwa,hampali2022keypoint} (Fig.~\ref{fig:transnet_compare_prev} (a)), which results in generalization failure due to the appearance gap between MoCap and ITW datasets.

We present InterWild, a framework for 3D interacting hands mesh recovery in the wild.
For the first sub-problem, InterWild takes a cropped single-hand image regardless of whether two hands are interacting or not, as shown in Fig.~\ref{fig:shnet_compare_prev} (b).
In this way, the 2D scales of interacting hands are normalized to those of a single hand.
Such normalization brings all single and interacting hands to the shared 2D scale space; hence, large-scale single-hand data of ITW datasets can be much more helpful compared to the counterpart without the normalization.

For the second sub-problem, InterWild takes geometric features without images, as shown in Fig.~\ref{fig:transnet_compare_prev} (b).
In particular, the output of the second sub-problem (\textit{i.e.}, 3D relative translation) is fully supervised only for MoCap samples to prevent the 2D-based weak supervision of ITW samples from deteriorating the 3D relative translation.
The geometric features are invariant to appearances, such as colors and illuminations, which can reduce the huge appearance gap between MoCap and ITW datasets and bring samples from two datasets to a shared appearance-invariant space.
Therefore, although the estimated 3D relative translation is supervised only on MoCap datasets and is not supervised on ITW datasets, our InterWild produces robust 3D relative translations on ITW datasets.

We show that our InterWild produces highly robust 3D interacting hand meshes from ITW images.
As 3D interacting hands recovery in the wild is barely studied, we hope that ours can give useful insight into future works.
For the continual study, we released our codes and trained models.

Our contributions can be summarized as follows.
\begin{itemize}
\item We present InterWild, a framework for the 3D interacting hands recovery in the wild.
\item For the separate left and right 3D hands, InterWild takes a cropped single-hand image regardless of whether hands are interacting or not so that all hands are brought to a shared 2D scale space.
\item For the 3D relative translation between two hands, InterWild takes only geometric features, which are invariant to appearances.
\end{itemize}

\section{Related works}~\label{sec:related_works}

\begin{figure*}[t]
\begin{center}
\includegraphics[width=0.8\linewidth]{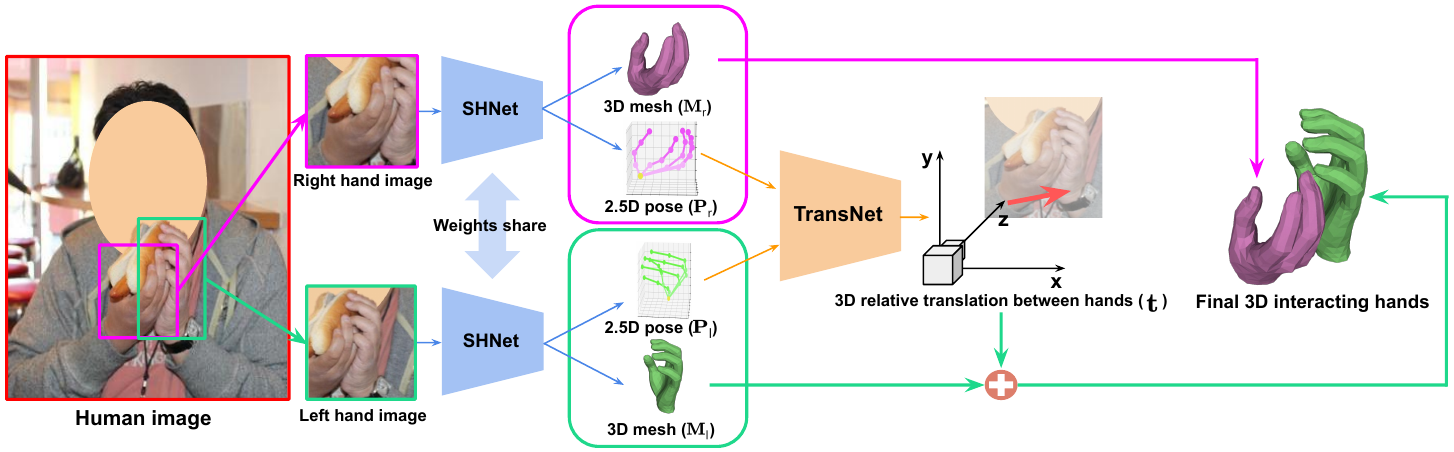}
\end{center}
\vspace*{-7mm}
\caption{
The overall pipeline of the proposed InterWild.
From the hand boxes, obtained by DetectNet, we crop and resize the hand area from the high-resolution human image.
Each right and left hand image is fed to SHNet, which produces 3D mesh and 2.5D heatmap.
Next, TransNet takes the 2.5D heatmap of two hands to produce the 3D relative translation between two hands.
Final 3D interacting hand meshes are obtained by adding the 3D relative translation to the left hand mesh.
For simplicity, we do not visualize DetectNet.
}
\vspace*{-3mm}
\label{fig:overall_pipeline}
\end{figure*}

\noindent\textbf{3D interacting hands recovery.}
Most of early works~\cite{oikonomidis2012tracking,ballan2012motion,tzionas2016capturing,taylor2016efficient,mueller2019real,wang2020rgb2hands} fit 3D hand models to geometric evidence, such as RGBD sequence~\cite{oikonomidis2012tracking}, hand segmentation map~\cite{mueller2019real}, and dense matching map~\cite{wang2020rgb2hands}.
Recently, Moon~\etal~\cite{moon2020interhand2,moon2022neuralannot} presented IH2.6M dataset, the first large-scale real-captured dataset that contains accurate GT 3D poses and meshes of interacting hands and a regression-based baseline model, InterNet.
Motivated by IH2.6M and InterNet, several regression-based methods have been proposed, which perform better than the above fitting-based methods.
Rong~\etal~\cite{rong2021monocular} proposed a two-stage framework to minimize collisions between two hands.
Zhang~\etal~\cite{zhang2021interacting} proposed a cascaded 3D interacting hand mesh estimation network, which sequentially refines 3D interacting hands.
Kwon~\etal~\cite{kwon2021h2o} presented a baseline for recovering 3D meshes of two hands while interacting with objects.
However, their system mostly focuses on the interaction between hands and objects, not between two hands.
Li~\etal~\cite{li2022interacting} proposed a graph convolutional network for accurate 3D interacting hand reconstruction.
Hampali~\etal~\cite{hampali2022keypoint} proposed a Transformer~\cite{vaswani2017attention}-based system that separates localization and identification of hand keypoints.
Di~\etal~\cite{di2022lwa} presented a lightweight system for 3D interacting hand mesh recovery.
In addition, there are several works~\cite{fan2021learning,kim2021end,meng2022hdr} that recover only 3D hand joint locations without 3D meshes.

The above 3D interacting hand reconstruction methods~\cite{moon2020interhand2,rong2021monocular,zhang2021interacting,li2022interacting,fan2021learning,kim2021end,meng2022hdr} fail to produce robust results on ITW datasets, while ours can.
There are two big differences.
First, they take a two-hand image as an input when two hands are interacting (Fig.~\ref{fig:shnet_compare_prev} (a)).
On the other hand, ours take a single-hand image regardless of whether two hands are interacting (Fig.~\ref{fig:shnet_compare_prev} (b)); hence, inputs are brought to the shared 2D scale space.
Second, they estimate 3D relative translation between two hands using an image with the 2D-based weak supervision (Fig.~\ref{fig:transnet_compare_prev} (a)).
On the other hand, ours estimates the relative translation only from geometric features (Fig.~\ref{fig:transnet_compare_prev} (b)), which are invariant to appearances, without the 2D-based weak supervision.

\noindent\textbf{Reducing appearance gap with geometric features.}
Several 3D body and hand mesh estimation methods have used geometric features to reduce the appearance gap between MoCap and ITW datasets.
Pose2Mesh~\cite{choi2020p2m} and Song~\etal~\cite{song2020human} take 2D body joint coordinates as an input to predict a 3D human body mesh.
Zhou~\etal~\cite{zhou2020monocular} predict a 3D single-hand mesh from 3D single-hand joint coordinates.
Zhang~\etal~\cite{zhang2020learning} utilizes body part UVI map for the 3D body mesh recovery.
Our InterWild is the first work that estimates robust 3D relative translation between two hands from geometric features.

\section{InterWild}
Fig.~\ref{fig:overall_pipeline} shows the overall pipeline of our InterWild, which consists of DetectNet, SHNet, and TransNet.
DetectNet detects hands from the input image.
Then, SHNet, a network for a single hand, takes each detected hand image as an input and outputs 3D mesh and 2.5D pose of each hand.
The 2.5D poses of the right and left hands are passed to TransNet, which outputs 3D relative translation between two hands.
The final 3D interacting hands are obtained by adding the 3D relative translation to the 3D mesh of the left hand.
DetectNet and SHNet follow architectures of Pose2Pose~\cite{moon2022hand4whole}.
Please refer to the supplementary material for their detailed architectures.

\begin{figure}[t]
\begin{center}
\includegraphics[width=0.7\linewidth]{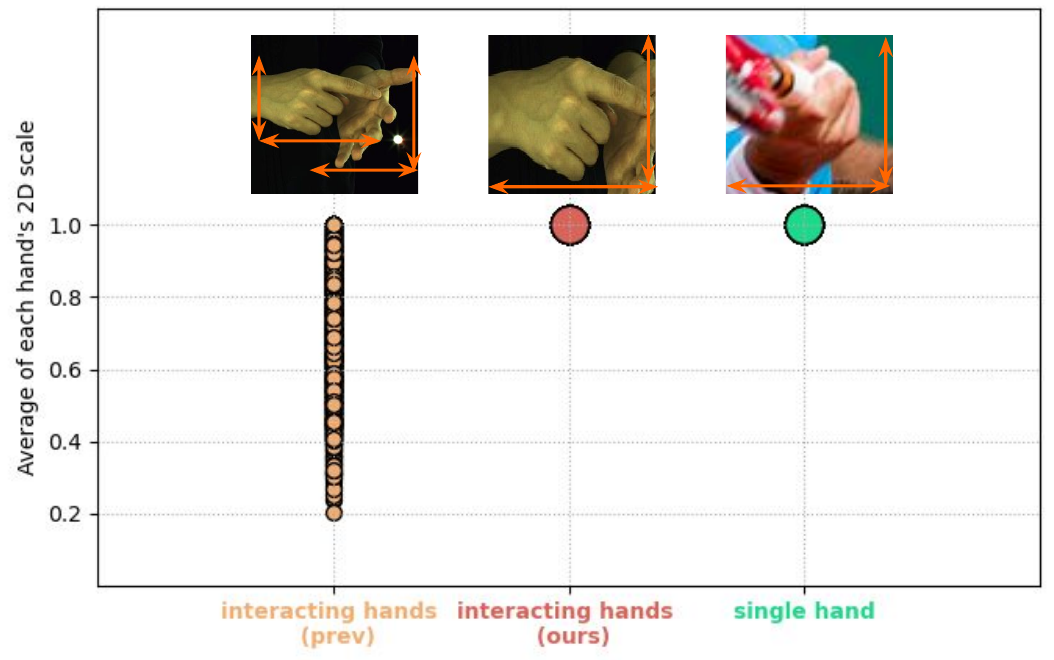}
\end{center}
\vspace*{-7mm}
\caption{
Average of each hand's width and height, where the width and height are normalized with the size of the input image.
We extended all hands' boxes to set their aspect ratio to 1 before calculating the scales.
\textcolor[RGB]{242,174,114}{Yellow}: Each hand in the two-hand image of IH2.6M~\cite{moon2020interhand2} when hands are interacting (Previous approach).
\textcolor[RGB]{217,100,89}{Brown}: Each hand in the single-hand image of IH2.6M~\cite{moon2020interhand2} when hands are interacting (Ours).
\textcolor[RGB]{30,216,139}{Green}: Each hand in the single-hand image of MSCOCO~\cite{lin2014microsoft}. 
}
\vspace*{-3mm}
\label{fig:sh_ih_dist_compare}
\end{figure}

\subsection{SHNet}

\noindent\textbf{Input: an image of a single hand.}
SHNet takes a single-hand image regardless of whether two hands are interacting or not, while previous methods take images with two hands when hands are interacting, as shown in Fig.~\ref{fig:shnet_compare_prev}.
Hence, 2D scales of interacting hands are normalized to those of a single hand.
Fig.~\ref{fig:sh_ih_dist_compare} shows that when we crop images to contain two hands when hands are interacting (\textit{i.e.}, previous methods. \textcolor[RGB]{242,174,114}{Yellow} in the figure), 2D scales of each hand in two-hand images have a very different distribution compared to those of each hand in single-hand images (\textcolor[RGB]{30,216,139}{Green} in the figure).
On the other hand, when we crop images to contain a single hand regardless of whether hands are interacting (\textit{i.e.}, ours. \textcolor[RGB]{217,100,89}{Brown} in the figure), 2D scales of each hand in two-hand images have almost the same distribution compared to those of each hand in single-hand images (\textcolor[RGB]{30,216,139}{Green} in the figure).
Such an analysis justifies our design of SHNet to take a single cropped hand.

The single-hand image is cropped and resized from the high-resolution human image using predicted boxes from the DetectNet.
Before cropping the hands, we double the width and height of boxes to prevent hands from missing and provide more surrounding context to SHNet.
The left hand image is horizontally flipped to the right hand; therefore, the input image always represents a right hand image.
The right hand and flipped left hand images are concatenated in the batch dimension and processed in a parallel way by the SHNet.
By taking the right hand and flipped left hand images, SHNet can focus only on learning to process right hand images, which can relieve the burdens of learning to process both right and left hand images.
Also, such flipping is helpful when the two hands are severely interacting so that boxes of two hands are largely overlapped.
For example, let us imagine that most of the left hand is occluded by the right hand.
Then, images from the left and right hand boxes would contain almost the same right hand.
By flipping the left hand image, the right hand in the original left hand image changes to the left hand.
We train SHNet to ignore the left hand in the input image and produce a 3D hand mesh of only the right hand in the input image.
Therefore, the output from the flipped left hand image is a 3D hand mesh of the occluded right hand, which is originally the occluded left hand.
The effectiveness of this flipping is shown in the experimental section.

\noindent\textbf{Output: 3D mesh and 2.5D pose of each hand.}
Using the network architecture of Pose2Pose~\cite{moon2022hand4whole}, our SHNet outputs 3D mesh and 2.5D pose~\cite{sun2018integral} of each hand.
We flip back the outputs of the flipped left hand image.
We denote the 3D mesh of left and right hands by $\mathbf{M}_\text{l}$ and $\mathbf{M}_\text{r}$, respectively.
Each 3D mesh is obtained by forwarding the estimated pose and shape parameters to a MANO~\cite{romero2017embodied} layer.
We subtract 3D meshes from their 3D root joint locations so that the 3D meshes are in the root joint-relative space.
In addition, we denote the 2.5D pose of left and right hands by $\mathbf{P}_\text{l} \in \mathbb{R}^{J \times 3}$ and $\mathbf{P}_\text{r} \in \mathbb{R}^{J \times 3}$, respectively.
$J$ indicates the number of single-hand joints.
The 2.5D pose encodes hand joint locations in 2.5D space.
The $x$- and $y$-axis of the $j$th 2.5D pose represent pixel coordinates of the $j$th joint, where the pixel space is defined in the input image of SHNet (\textit{i.e.}, single-hand image).
The $z$-axis is defined in the root joint-relative depth space.

\subsection{TransNet}~\label{subsec:transnet}
\begin{figure}[t]
\begin{center}
\includegraphics[width=\linewidth]{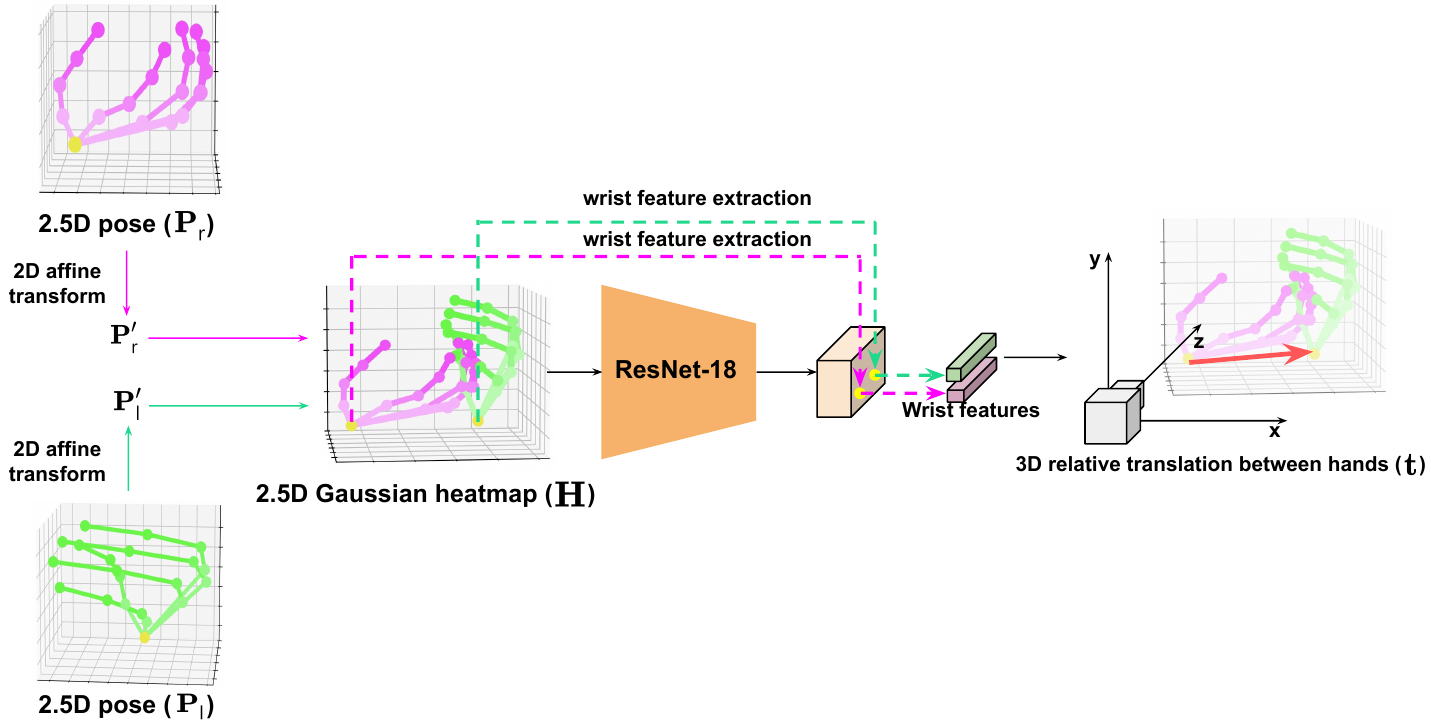}
\end{center}
\vspace*{-7mm}
\caption{
The overall pipeline of TransNet.
It applies a 2D affine transformation to the 2.5D pose of each left and right hand to bring them to the union hand box space of the original input image.
Then, wrist features are extracted for the 3D relative translation estimation.
}
\vspace*{-3mm}
\label{fig:transnet}
\end{figure}

Fig.~\ref{fig:transnet} shows the overall pipeline of TransNet, a network to predict 3D relative translation between two hands.

\noindent\textbf{Input: 2.5D poses of two hands.}
TransNet takes 2.5D poses of two hands, while previous methods take images with two hands, as shown in Fig.~\ref{fig:transnet_compare_prev}.
The 2.5D poses of two hands are from SHNet, which are denoted by $\mathbf{P}_\text{r}$ and $\mathbf{P}_\text{l}$.
Before forwarding them, we apply 2D affine transformations to $\mathbf{P}_\text{r}$ and $\mathbf{P}_\text{l}$, which transform the input space of SHNet (\textit{i.e.}, an image of a single hand) to a union of two-hand boxes space (\textit{i.e.}, an image of two hands).
By warping them to the union hand box space, we can get a relative 2D scale and translation between two hands in the 2D pixel space.
Based on such relative 2D information and pose information, TransNet predicts the 3D relative translation.

For example, when $xy$ distance of two hands' 2.5D pose is small, $(x,y)$ of the 3D relative translation are close to zero.
Also, when one hand takes smaller area in input $xy$ space, that hand might have larger depth; however, not always true as hands are deformable.
When a hand is in neutral pose and the other one is in fist pose, their 3D relative depth can be zero although the hand with fist pose takes smaller area.
Hence, pose is necessary to determine $z$ of the 3D relative translation.
Please note that the 2D affine transformations do not affect the depths of each 2.5D pose; hence, the depths of each 2.5D pose still represent the root joint-relative depths of each hand.
We denote the transformed $\mathbf{P}_\text{r}$ and $\mathbf{P}_\text{l}$ by $\mathbf{P}_\text{r}'$ and $\mathbf{P}_\text{l}'$, respectively.

The 2.5D pose of the right hand $\mathbf{P}_\text{r}'$ and left hand $\mathbf{P}_\text{l}'$ are converted to 2.5D Gaussian heatmaps by making a Gaussian blob around the coordinates.
By converting coordinates to heatmaps, we can exploit the strong feature extraction power of ResNet~\cite{he2016deep} as ResNet takes tensor inputs, not vector inputs.
Then, we concatenate the 2.5D Gaussian heatmap of two hands in a channel dimension, denoted by $\mathbf{H} \in \mathbb{R}^{2J \times D \times H \times W}$.
$D$, $H$, and $W$ represent the depth, height, and width of the 2.5D heatmap, respectively, and we set them to 64.

\noindent\textbf{Output: 3D relative translation between two hands.}
We predict the 3D relative translation between two hands $\mathbf{t} \in \mathbb{R}^3$ from the 2.5D Gaussian heatmap $\mathbf{H}$.
We pass $\mathbf{H}$ to ResNet-18~\cite{he2016deep}, which produces a feature map $\mathbf{F} \in \mathbb{R}^{C \times H/8 \times W/8}$.
$C=512$ represents the channel dimension of $\mathbf{F}$.
We use the original ResNet-18 after dropping the first convolutional block to reduce the downsampling and the last fully-connected layers.
As the 3D relative translation represents a 3D relative location of the left wrist from the right wrist, extracting useful wrist information is a key for accurate 3D relative translation.
However, most existing methods~\cite{moon2020interhand2,rong2021monocular,fan2021learning,zhang2021interacting} perform global average pooling (GAP) to the last feature map of backbones and pass the output to several fully connected layers.
As GAP simply averages the spatial dimension, it might not be effective to capture useful wrist information.
Instead, we perform a bilinear interpolation at the 2D positions of left and right wrists in $\mathbf{F}$, where the 2D wrist positions are from $\mathbf{P}_\text{r}'$ and $\mathbf{P}_\text{l}'$.
The extracted wrist features are concatenated with the 2D wrist coordinates and fed to a linear layer, which finally produces 3D relative translation $\mathbf{t}$.
The effectiveness of our wrist feature extraction compared to previous GAP-based approaches is shown in the experimental section.

\subsection{Final outputs and loss functions}~\label{subsec:loss}
The final 3D interacting hand meshes consist of 1) the 3D mesh of the right hand $\mathbf{M}_\text{r}$ and 2) a summation of the 3D mesh of the left hand $\mathbf{M}_\text{l}$ and the 3D relative translation $\mathbf{t}$.
We train InterWild in an end-to-end manner by minimizing $L1$ distance between predicted and GT boxes, MANO parameters, 3D joint coordinates, and 3D relative translation.
Please note that the 3D relative translation is only supervised by MoCap datasets as ITW datasets do not provide GTs.

\section{Experiments}

\subsection{Datasets}

\noindent\textbf{Train sets.}
IH2.6M~\cite{moon2020interhand2} (H) and the whole-body version of MSOCO~\cite{lin2014microsoft,jin2020whole} are used for the training.
During the training, the mini-batch consists of half-IH2.6M and half-MSCOCO samples.

\noindent\textbf{Test sets.}
As our primary goal is 3D interacting hand mesh recovery \emph{in the wild}, we use Hands In Action dataset (HIC)~\cite{tzionas2016capturing} as our main test set.
HIC~\cite{tzionas2016capturing} contains single and interacting hand sequences captured with an RGBD camera.
HIC provides 3D GT meshes of hands~\cite{hasson2019learning}, fitted to a 3D point cloud.
Although HIC is captured in an indoor environment, it contains images with much more diverse and realistic appearances compared to those of IH2.6M.
Also, as we do not use HIC during the training, its appearances are not exposed to the network; hence, we believe the test performance of networks on HIC represents generalizability to unseen appearances, necessary for the 3D interacting hand mesh recovery in the wild.
Additionally, we report errors on IH2.6M (H) as it is one of the representative datasets for the 3D interacting hand mesh recovery although it is a MoCap dataset.
Qualitative results are shown on MSCOCO, which is the most widely used ITW dataset due to its diverse appearances.

\subsection{Evaluation metrics}

\noindent\textbf{MPJPE and MPVPE.}
Mean per-joint position error (MPJPE) and mean per-vertex position error (MPVPE) evaluate 3D joint and mesh vertex positions, respectively.
It represents the average 3D joint and mesh vertex distance (mm) between the predicted and GT, after aligning those with a root joint translation.
MPJPE and MPVPE are used to measure 3D errors of 3D mesh of each hand.

\noindent\textbf{MRRPE.}
Mean relative-root position error (MRRPE) evaluates 3D relative translation between two hands.
It calculates a 3D distance (mm) between the predicted and GT right hand root-relative left hand root position.

\subsection{Ablation study}

\begin{table}[t]
\footnotesize
\centering
\setlength\tabcolsep{1.0pt}
\def\arraystretch{1.1}
\begin{tabular}{C{4.0cm}|C{2.0cm}|C{2.0cm}}
\specialrule{.1em}{.05em}{.05em}
Inputs of SHNet & HIC~\cite{tzionas2016capturing} & IH2.6M~\cite{moon2020interhand2} \\ \hline
Two-hand image & 29.80 / 35.86 &  11.36 / 13.20 \\ 
\textbf{Single-hand image (Ours)} & \textbf{15.65} / \textbf{15.70} & \textbf{11.12} / \textbf{13.01} \\ 
\specialrule{.1em}{.05em}{.05em}
\end{tabular}
\vspace*{-3mm}
\caption{MPVPE comparisons between SHNets that take an image of 1) two hands and 2) a single hand as an input when hands are interacting.
Both settings take a single hand image when hands are not interacting.
The left and right numbers for each setting represent errors from single and interacting hand sequences, respectively.}
\label{table:shnet_input}
\end{table}

\noindent\textbf{SHNet: Effectiveness of taking an image of a single hand.}
Table~\ref{table:shnet_input} shows that when hands are interacting, taking a single hand image (Fig.~\ref{fig:shnet_compare_prev} (b)) produces lower SHNet's errors compared to taking a two-hands image (Fig.~\ref{fig:shnet_compare_prev} (a)), especially on HIC.
This shows that our approach to taking a single hand image when hands are interacting is especially helpful in the wild.
The reason for our superior result is that when SHNet takes a single hand image, large-scale 3D interacting hand data of MoCap dataset~\cite{moon2020interhand2} and large-scale 2D single hand data of in-the-wild dataset~\cite{lin2014microsoft} are brought to a shared 2D scale space.
On the other hand, when taking images of two hands when hands are interacting like previous works~\cite{moon2020interhand2,rong2021monocular,zhang2021interacting,li2022interacting,fan2021learning,kim2021end,meng2022hdr}, the 2D scale of each hand has a very different distribution, as shown in Fig.~\ref{fig:sh_ih_dist_compare}; hence, learning to process inputs with a very different distribution can be a burden to SHNet.
The small gap of IH2.6M is because such a burden can be relieved by large-scale interacting hand datasets, such as IH2.6M.
We observed that reducing the number of IH2.6M's interacting hand samples makes the same tendency of HIC, which supports our claim.
However, collecting large-scale interacting hand data in the wild is greatly challenging; even collecting 2D data requires a huge amount of effort due to the severe self-similarity and occlusions.
We effectively relieve such data collection burden by bringing the two datasets to a shared 2D scale space.
Both settings take a single hand when hands are not interacting, and the first setting takes a two-hand image when hands are interacting.
For the setting that takes an image of two hands, we doubled the output part of SHNet to estimate both left and right hands at the same time.

\begin{table}[t]
\footnotesize
\centering
\setlength\tabcolsep{1.0pt}
\def\arraystretch{1.1}
\begin{tabular}{C{4.5cm}|C{1.7cm}|C{1.7cm}}
\specialrule{.1em}{.05em}{.05em}
Settings & HIC~\cite{tzionas2016capturing} & IH2.6M~\cite{moon2020interhand2} \\ \hline
Without flip / shared weights & 15.80 / 17.36 & 12.54 / 18.21 \\ 
Without flip / separated weights & \textbf{15.34} / 16.96 & 11.48 / 13.77 \\ 
\textbf{With flip / shared weights (Ours)} & 15.65 / \textbf{15.70} & \textbf{11.12} / \textbf{13.01} \\ 
\specialrule{.1em}{.05em}{.05em}
\end{tabular}
\vspace*{-3mm}
\caption{MPVPE comparisons of SHNets that take 1) left and right hands and 2) flipped left hand and right hand.
`Shared weights' indicates that SHNet's weights are shared for left and right hand inputs, while `separated weights' indicates that the weights are not shared. 
The left and right numbers for each setting represent errors from single and interacting hand sequences, respectively.}
\label{table:shnet_flip}
\end{table}

\begin{figure}[t]
\begin{center}
\includegraphics[width=\linewidth]{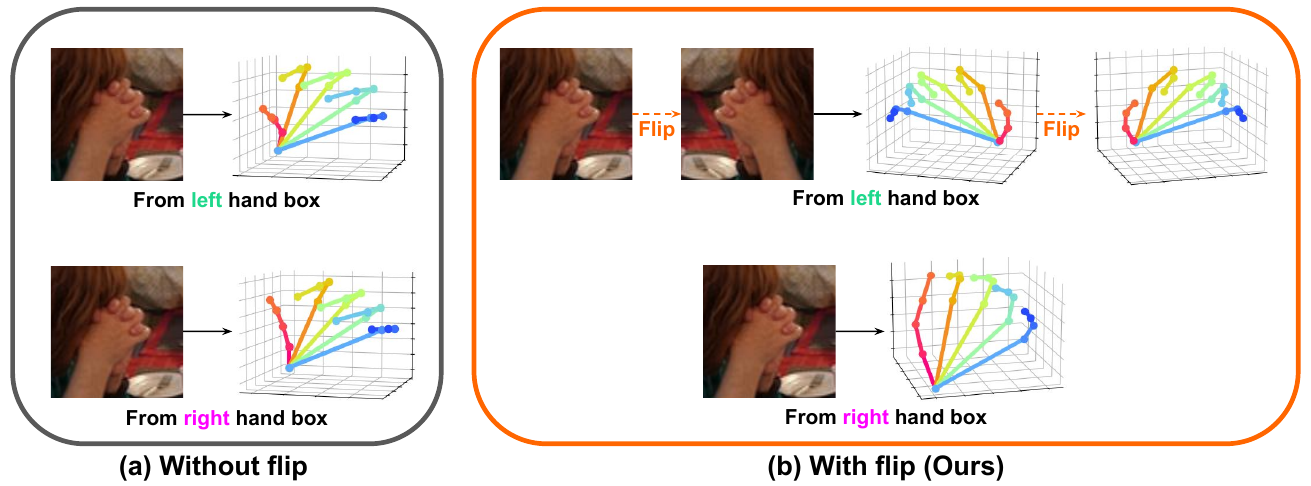}
\end{center}
\vspace*{-7mm}
\caption{
Comparison between 2.5D hand pose from SHNets that take (a) an image of left and right hand images and (b) an image of \emph{flipped left hand} and right hand.
The result from the flipped left hand is flipped back.
}
\vspace*{-3mm}
\label{fig:ablation_shnet_flip}
\end{figure}

\noindent\textbf{SHNet: Effectiveness of flipping left hand images.}
Table~\ref{table:shnet_flip} shows that flipping the left hand is necessary for SHNet's accurate results when hands are interacting.
Without the flipping, SHNet can take almost the same single hand image when hands are severely overlapped, as shown in Fig.~\ref{fig:ablation_shnet_flip} (a).
As input images are almost the same, SHNet outputs almost the same two 3D hands for both left and right hand images.
On the other hand, when we pass a flipped left hand image, SHNet can differentiate the two images.
As SHNet is trained to ignore all left hands and recover only right hands, this can be seen as an \emph{implicit de-occlusion} of left hands from the input image.
Fig.~\ref{fig:ablation_shnet_flip} (b) shows that even when two hands are severely overlapped, our SHNet can recover correct 3D hands.

Instead of flipping, one can double the output part of SHNet and train SHNet to output left and right hands at the same time.
In this way, when the input hand images are almost the same due to the severe overlap, the output part of SHNet is trained to distinguish left and right hands.
However, we observed that our flipping results in better results than this variant.
We think this is because, in our setting, the handedness of the input image is normalized to the right hand, while the variant is not.
Such normalization can relieve the burden of SHNet.
Please note that a combination of `with flip' and `separated weights' is impossible as flipping normalizes the handedness.

\begin{table}[t]
\scriptsize
\centering
\setlength\tabcolsep{1.0pt}
\def\arraystretch{1.1}
\begin{tabular}{C{3.0cm}C{1.5cm}|C{1.5cm}|C{1.7cm}}
\specialrule{.1em}{.05em}{.05em}
Inputs of TransNet & weak sup. & HIC~\cite{tzionas2016capturing} & IH2.6M~\cite{moon2020interhand2} \\ \hline
\multirow{2}{*}{Img.} & \xmark & 206.83 &  27.67 \\ 
& \cmark & 215.35 & 35.72 \\ \hline
\multirow{2}{*}{Img. + 2.5D hm.} & \xmark & 54.36 & \textbf{27.19} \\
& \cmark & 58.53 & 33.15 \\ \hline
\multirow{2}{*}{2D hm.} & \xmark& 38.64 & 31.51 \\
& \cmark & 51.19 & 35.51 \\ \hline
\multirow{2}{*}{2.5D hm.} & \xmark \textbf{(Ours)} & \textbf{31.35} &  29.29 \\ 
& \cmark & 61.05 & 33.91 \\
\specialrule{.1em}{.05em}{.05em}
\end{tabular}
\vspace*{-3mm}
\caption{MRRPE comparisons of TransNets that take various inputs and are trained without and with the 2D-based weak supervision. The hm. represents a heatmap.}
\label{table:transnet_input_weak_sup}
\end{table}

\begin{table}[t]
\footnotesize
\centering
\setlength\tabcolsep{1.0pt}
\def\arraystretch{1.1}
\begin{tabular}{C{3.5cm}|C{2.0cm}|C{2.0cm}}
\specialrule{.1em}{.05em}{.05em}
Settings & HIC~\cite{tzionas2016capturing} & IH2.6M~\cite{moon2020interhand2} \\ \hline
GAP & 39.85 & \textbf{29.14}  \\ 
All joint features & 48.99 & 31.57 \\
\textbf{Wrist features (Ours)} & \textbf{31.35} &  29.29 \\ 
\specialrule{.1em}{.05em}{.05em}
\end{tabular}
\vspace*{-3mm}
\caption{MRRPE comparisons between TransNet that outputs the 3D relative translation with various feature extraction settings.}
\vspace*{-3mm}
\label{table:transnet_wrist_features}
\end{table}

\noindent\textbf{TransNet: Effectiveness of the geometric inputs.}
Table~\ref{table:transnet_input_weak_sup} shows that taking 2.5D heatmaps as an input of TransNet (Fig.~\ref{fig:transnet_compare_prev} (b)) achieves the lowest MRRPE on HIC and comparable results on IH2.6M.
Our 2.5D heatmap achieves better results than the 2D heatmap due to the additional depth information of each hand.
An interesting result is that using an image as an input (the first and second rows) achieves good results on IH2.6M, but bad results on HIC.
Such a setting with the image input is similar to previous methods~\cite{moon2020interhand2,rong2021monocular,zhang2021interacting,fan2021learning,li2022interacting}, while IntagHand~\cite{li2022interacting} additionally uses segmentation and DensePose.

There are two reasons for this setting's high MRRPE on HIC.
First, when the 2D-based weak supervision is disabled, the huge appearance gap between ITW and IH2.6M is the main reason.
Without the weak supervision, TransNet is supervised only on IH2.6M.
Images of MoCap datasets, including IH2.6M, have monotonous colors with artificial illuminations, which are far from those of ITW images.
On the other hand, we use pure geometric features (\textit{i.e.}, 2.5D heatmap) as it is invariant to appearances.
Thanks to the invariance, our TransNet successfully generalizes to ITW datasets although it is trained only on IH2.6M.
Second, when the 2D-based weak supervision is enabled, the weak supervision deteriorates the relative translation due to the scale ambiguity of the relative translation.
A detailed analysis of the 2D-based weak supervision is provided below.

\noindent\textbf{TransNet: Bad effect of the 2D-based weak supervision.}
Table~\ref{table:transnet_input_weak_sup} shows that introducing the 2D-based weak supervision for the estimation of the 3D relative translation between two hands, similar to IntagHand~\cite{li2022interacting}, deteriorates MRRPE for all inputs of TransNet and for both evaluation benchmarks.
This is because, unlike the 3D scales of hands that are strongly constrained with the shape parameter of MANO, 3D scales of the 3D relative translation are very weakly constrained.
For example, we can put two hands near or far based on how they are interacting with each other, while the size of adults' hands is usually around 15 cm.
Without such a strong constraint, the 3D relative translation can be an arbitrary value due to the wrong 3D global translation.
Fig.~\ref{fig:motivation} (b) shows that when the 3D global translation is wrong (\textcolor{orange}{\textbf{\textcircled{\raisebox{-0.9pt}{1}}}} in the figure), the 3D relative translation is supervised to be wrong one (\textcolor{orange}{\textbf{the orange arrow}} in the figure).
Please refer to the supplementary material for how we introduced the 2D-based weak supervision.

\begin{table}[t]
\footnotesize
\centering
\setlength\tabcolsep{1.0pt}
\def\arraystretch{1.1}
\begin{tabular}{C{2.4cm}|C{1.7cm}C{1.1cm}|C{1.7cm}C{1.1cm}}
\specialrule{.1em}{.05em}{.05em}
\multirow{2}{*}{Methods} & \multicolumn{2}{c|}{HIC~\cite{tzionas2016capturing}} & \multicolumn{2}{c}{IH2.6M~\cite{moon2020interhand2}} \\
& MPVPE & MRRPE & MPVPE & MRRPE \\ \hline
IHMR~\cite{rong2021monocular} & 30.76 / 46.38 & 119.64 & 15.35 / 18.53 & 33.39 \\
Zhang~\etal~\cite{zhang2021interacting} & 23.53 / 31.79 & 110.25 & 11.76 / 14.17 & 31.56 \\ 
IntagHand~\cite{li2022interacting} & 18.83 / 27.31 & 52.46 & 11.18 / 13.49 & 29.31 \\
\textbf{InterWild (Ours)} & \textbf{15.65} / \textbf{15.70} & \textbf{31.35} & \textbf{11.12} / \textbf{13.01} & \textbf{29.29} \\ \specialrule{.1em}{.05em}{.05em}
\end{tabular}
\vspace*{-3mm}
\caption{Comparison of our InterWild and 3D interacting hand mesh estimation methods.
The left and right numbers for each setting represent errors from single and interacting hand sequences, respectively.
}
\vspace*{-3mm}
\label{table:comparison_sota_mesh}
\end{table}

\begin{table}[t]
\footnotesize
\centering
\setlength\tabcolsep{1.0pt}
\def\arraystretch{1.1}
\begin{tabular}{C{2.5cm}|C{2.0cm}C{1.5cm}}
\specialrule{.1em}{.05em}{.05em}
Methods & MPJPE & MRRPE  \\ \hline
AIH~\cite{meng2022hdr} &  76.83 / 36.05 & N/A \\
\textbf{InterWild (Ours)} & \textbf{16.00} / \textbf{16.17} & \textbf{31.35} \\ \specialrule{.1em}{.05em}{.05em}
\end{tabular}
\vspace*{-3mm}
\caption{Comparison of our InterWild and 3D interacting hand pose estimation methods on HIC.
The left and right numbers for each setting represent errors from single and interacting hand sequences, respectively.
}
\vspace*{-5mm}
\label{table:comparison_sota_pose}
\end{table}

\begin{figure}[t]
\begin{center}
\includegraphics[width=\linewidth]{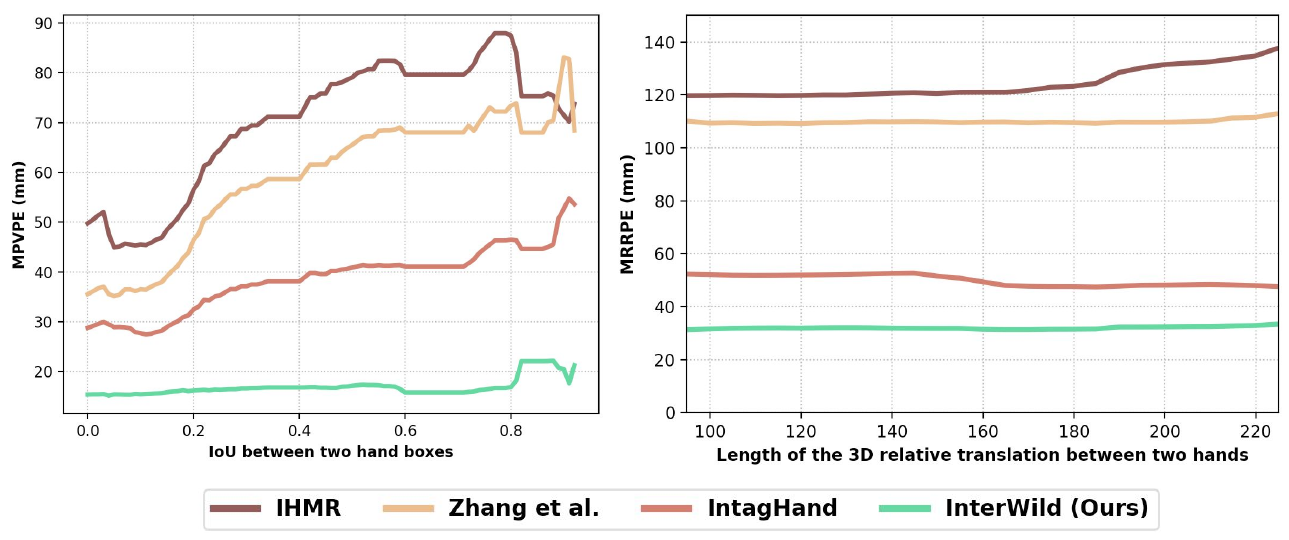}
\end{center}
\vspace*{-7mm}
\caption{
Comparison with previous methods~\cite{rong2021monocular,zhang2021interacting,li2022interacting} on HIC.
For each $x$-axis value, denoted by $\tau$, MPVPE and MRRPE of $y$-axis are calculated from samples whose $x$-axis values are larger than $\tau$.
}
\vspace*{-3mm}
\label{fig:mpvpe_mrrpe_hic}
\end{figure}

\begin{figure*}[t]
\begin{center}
\includegraphics[width=0.77\linewidth]{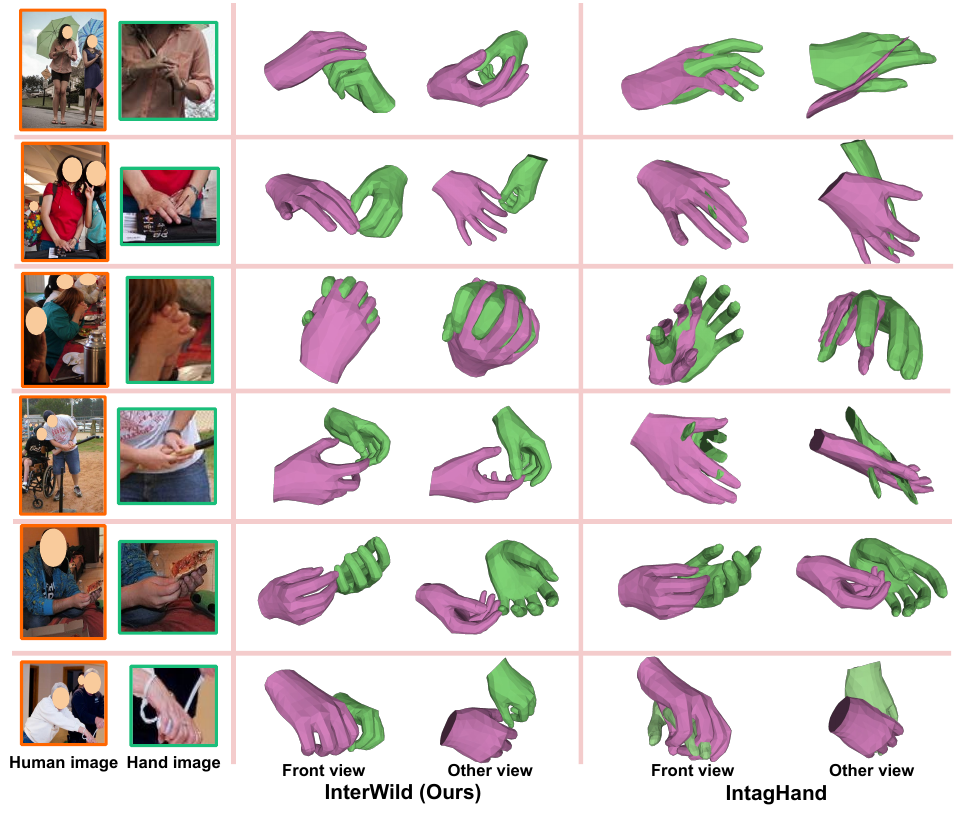}
\end{center}
\vspace*{-7mm}
\caption{
Qualitative comparison between our InterWild and IntagHand~\cite{li2022interacting} on MSCOCO validation set.
Ours detects hand boxes from the human images, while IntagHand takes the hand images using GT boxes.
}
\vspace*{-5mm}
\label{fig:qualitative_comparison}
\end{figure*}

\noindent\textbf{TransNet: Effectiveness of the wrist features.}
Table~\ref{table:transnet_wrist_features} shows that our wrist feature-based estimation of 3D relative translation achieves better results than the previous widely used GAP~\cite{moon2020interhand2,rong2021monocular,zhang2021interacting,fan2021learning}.
This is because GAP averages the entire spatial domain; therefore, its output is not guaranteed to contain essential wrist information, necessary for the 3D relative translation between two hands.
On the other hand, we explicitly extract wrist features, which produces better results.
Interestingly, using features of all joints performs worst.
We think that this is because features from all joints contain too much unnecessary information, which gives a burden to the regressor.

\subsection{Comparison with state-of-the-art methods}
Table~\ref{table:comparison_sota_mesh} and Fig.~\ref{fig:mpvpe_mrrpe_hic} show that ours outperforms all existing 3D interacting hand mesh recovery methods on HIC and IH2.6M datasets.
It is noteworthy that the MPVPE and MRRPE gap is especially large on HIC, which shows the robustness of InterWild to ITW environments.
This is because of our domain-sharing approach, depicted in Fig.~\ref{fig:shnet_compare_prev} and ~\ref{fig:transnet_compare_prev}.
Table~\ref{table:comparison_sota_pose} additionally demonstrates the superiority of our InterWild.
AIH~\cite{meng2022hdr} used additional synthetic datasets for the robust results on unseen appearances.
However, such synthetic datasets are still built on top of the IH2.6M dataset, which has a severe appearance gap from that of ITW datasets.
On the other hand, ours effectively reduces the domain gap by bringing inputs of SHNet and TransNet to shared domains, which results in a strong performance on ITW datasets.
Finally, Fig.~\ref{fig:qualitative_comparison} visually demonstrates that ours successfully recovers 3D meshes from ITW images, while previous state-of-the-art method~\cite{li2022interacting} fails to.

Publicly released models of previous works in Table~\ref{table:comparison_sota_mesh} are trained only on IH2.6M, and their training codes are not available.
Therefore, we reproduced their networks based on their testing codes and re-trained all of their networks on IH2.6M and MSCOCO like ours for a fair comparison.
We will verify our reproduce results in the supplementary material.
For the evaluation, we do not align the scale with GTs following Moon~\etal~\cite{moon2020interhand2}.
We found that Keypoint Transformer~\cite{hampali2022keypoint} produces bad results when MSCOCO is incorporated in the training set as it requires 3D GTs, which does not exist in MSCOCO, for the camera parameter loss function.
All previous methods use GT hand boxes during the inference following their settings, while ours uses predicted hand boxes from DetectNet.

\section{Conclusion}

We present InterWild, a framework for the 3D interacting hand mesh recovery in the wild.
InterWild effectively reduces the domain gap between MoCap and ITW datasets.
To this end, it takes an image of a single hand regardless of whether hands are interacting for the estimation of 3D meshes of left and right hands.
In addition, it takes geometric features for the estimation of 3D relative translation between two hands.
Integrating our InterWild with whole-body 3D human mesh estimation methods can be a promising future research direction.


\begin{center}
\textbf{\large Supplementary Material for \\ ``Bringing Inputs to Shared Domains for \\ 3D Interacting Hands Recovery in the Wild"}
\end{center}

\setcounter{section}{0}
\setcounter{table}{0}
\setcounter{figure}{0}

\renewcommand{\thesection}{\Alph{section}}   
\renewcommand{\thetable}{\Alph{table}}   
\renewcommand{\thefigure}{\Alph{figure}}

In this supplementary material, we provide more experiments, discussions, and other details that could not be included in the main text due to the lack of pages.
The contents are summarized below:
\begin{enumerate}[nosep, label=\Alph*.]  
    \item Qualitative comparison
    \item Ablation study on the architecture of TransNet
    \item Verification of reproduced results in Table 5
    \item Clarification of 2D-based weak supervision in Table 3
    \item Architecture of DetectNet
    \item Architecture of SHNet
    \item Implementation details
    \item Limitations
\end{enumerate}

\begin{figure*}[t]
\begin{center}
\includegraphics[width=\linewidth]{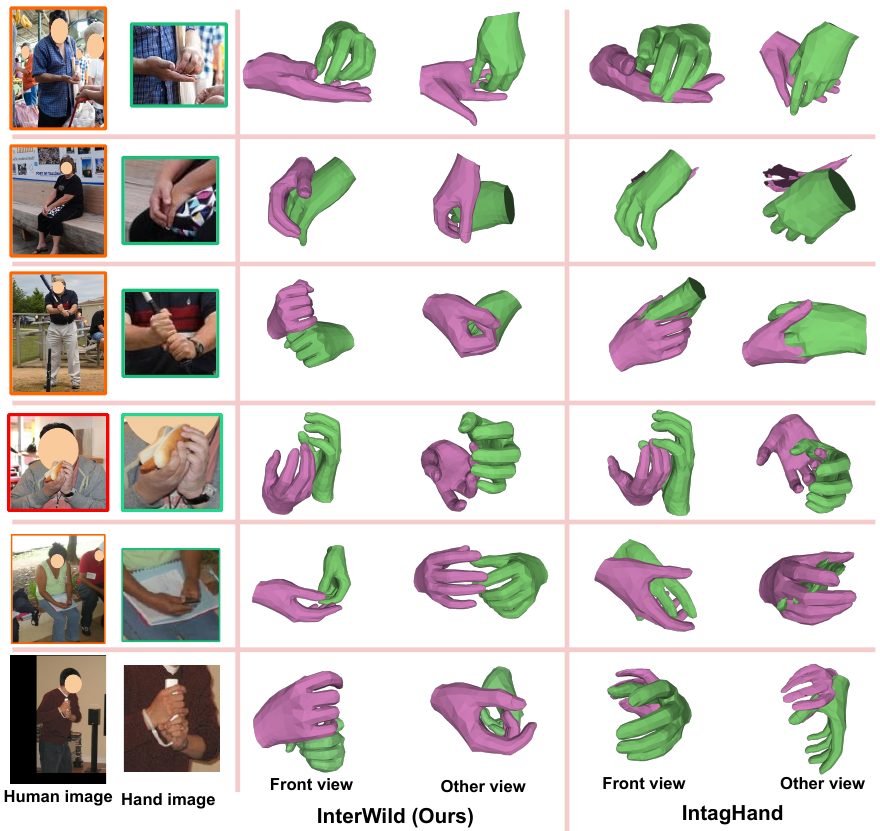}
\end{center}
\vspace*{-5mm}
\caption{
Qualitative comparison between our InterWild and IntagHand~\cite{li2022interacting} on MSCOCO.
Ours detects hand boxes from the human images, while IntagHand takes the hand images using GT boxes.
}
\label{fig:qualitative_comparison_2}
\end{figure*}

\section{Qualitative comparisons}

Fig.~\ref{fig:qualitative_comparison_2} shows that ours produces much more robust results than IntagHand~\cite{li2022interacting} on in-the-wild images.
Overall, IntagHand produces reasonable 3D hand mesh for a visible hand but fails to recover the other occluded hand.
Also, it suffers from depth ambiguity as the first and fourth rows show, where the 2D error is small but the 3D error is large.
The fourth row also shows that IntagHand fails to recover 3D relative translation between two hands due to the depth ambiguity, while ours successfully recovers.
Finally, we think the reason why IntagHand produces non-hand shape meshes is that IntagHand directly regresses the 3D coordinates of 3D hand meshes.
On the other hand, ours regresses MANO parameters and 3D meshes are obtained by forwarding the parameters to the MANO layer.
\section{Ablation study on the architecture of TransNet}

Table~\ref{table:transnet_architecture} shows that our fully convolutional network (FCN)-based architecture performs the best compared to widely used fully-connected (FC) network architecture and Transformer~\cite{vaswani2017attention}.
We think this is because our FCN can explicitly utilize the spatial relationship between voxels using the 2.5D heatmap representation.
On the other hand, FC-based and Transformer-based settings take 2.5D coordinates as input.
This result is consistent with previous studies~\cite{moon2020i2l}, which demonstrates the superiority of heatmap representation over coordinate representation.
As proposing better network architecture for TransNet is not our main focus, we believe developing its network architecture can be an interesting future direction.
We used the architecture of Martinez~\etal~\cite{martinez2017simple} for the FC setting, and Zheng~\etal~\cite{zheng20213d} for the Transformer setting.

\begin{table}[t]
\small
\centering
\setlength\tabcolsep{1.0pt}
\def\arraystretch{1.1}
\begin{tabular}{C{3.5cm}|C{2.0cm}|C{2.0cm}}
\specialrule{.1em}{.05em}{.05em}
Settings & HIC~\cite{tzionas2016capturing} & IH2.6M~\cite{moon2020interhand2} \\ \hline
FC & 56.76 & 29.85 \\ 
Transformer & 64.49 & 33.59 \\
\textbf{FCN (Ours)} & \textbf{31.35} & \textbf{29.29} \\ 
\specialrule{.1em}{.05em}{.05em}
\end{tabular}
\vspace*{-3mm}
\caption{MRRPE comparisons between TransNet that have various architectures.}
\label{table:transnet_architecture}
\end{table}

\begin{table}[t]
\small
\centering
\setlength\tabcolsep{1.0pt}
\def\arraystretch{1.1}
\begin{tabular}{C{2.0cm}C{2.2cm}C{2.2cm}|C{1.2cm}}
\specialrule{.1em}{.05em}{.05em}
Settings & Translation align & Scale align & MPVPE \\ \hline
Official model & Middle root & \cmark & 9.36 \\ 
Reproduced & Middle root & \cmark & 9.34 \\ \hline
Official model & Middle root & \xmark & 9.99 \\ 
Reproduced & Middle root & \xmark & 9.81 \\ \hline
Official model & Wrist & \xmark & 15.24 \\ 
Reproduced & Wrist & \xmark & 14.12 \\
\specialrule{.1em}{.05em}{.05em}
\end{tabular}
\vspace*{-3mm}
\caption{Reproduce verification of IntagHand~\cite{li2022interacting}.
We report MPVPE on interacting hand sequences of InterHand2.6M.
}
\label{table:intaghand_reproduce}
\end{table}

\begin{table}[t]
\small
\centering
\setlength\tabcolsep{1.0pt}
\def\arraystretch{1.1}
\begin{tabular}{C{2.0cm}C{2.2cm}|C{1.2cm}}
\specialrule{.1em}{.05em}{.05em}
Settings & Scale align & MPVPE \\ \hline
Original & \cmark & 10.40 \\ 
Reproduced & \cmark & 10.70 \\
Reproduced & \xmark & 14.20 \\
\specialrule{.1em}{.05em}{.05em}
\end{tabular}
\vspace*{-3mm}
\caption{Reproduce verification of Zhang~\etal~\cite{zhang2021interacting}.
We report MPVPE on interacting hand sequences of InterHand2.6M.
}
\label{table:zhang_reproduce}
\end{table}

\section{Verification of reproduced results in Table~\ref{table:comparison_sota_mesh}}

Table~\ref{table:intaghand_reproduce} and ~\ref{table:zhang_reproduce} verify our reproduce results of IntagHand~\cite{li2022interacting} and Zhang~\etal~\cite{zhang2021interacting} on InterHand2.6M, respectively.
For the verification of IntagHand~\cite{li2022interacting}, we used their officially released pre-trained model and evaluation code.
As their model is trained only on interacting hand sequences of InterHand2.6M, we also trained a model following their training set for the verification.
Their original evaluation setting is the first row of Table~\ref{table:intaghand_reproduce}.
On the other hand, our evaluation setting in Table~\ref{table:comparison_sota_mesh} of the main manuscript is the last row as most 3D hand recovery works~\cite{moon2020interhand2,boukhayma20193d,choi2020p2m,Freihand2019,moon2020i2l} align translation with the wrist joint, a \emph{root joint} in hand kinematic space.
Using the middle root joint reduces the errors as the length of the kinematic chain to the fingertips becomes much shorter.
Please note that the reproduced numbers can be different from those of Table~\ref{table:comparison_sota_mesh} in the main manuscript as the results in Table 5 are from additional training on MSCOCO.

For the verification of Zhang~\etal~\cite{zhang2021interacting}, we take numbers from their paper as their released pre-trained model does not work.
Following their training protocol, we trained a model only on the interacting hand sequence of InterHand2.6M.
Please note that the reproduced numbers can be different from those of Table~\ref{table:comparison_sota_mesh} in the main manuscript as the results in Table 5 are from additional training on MSCOCO.

\section{2D-based weak supervision in Table~\ref{table:transnet_input_weak_sup}}
We describe how the 2D-based weak supervision in Table~\ref{table:transnet_input_weak_sup} of the main manuscript is introduced.
We modified TransNet to output 3D global translation of the right hand and 3D relative translation between two hands from the two hand input.
We observed this produces better results than estimating 3D global translations of both hands.
The 3D global translation of the left hand is obtained by adding the 3D relative translation to the 3D global translation of the right hand.
Then, we translate root joint-relative 3D joint coordinates of each hand (\textit{i.e.}, output of SHNet) using the 3D global translation of each hand.
The translated 3D joint coordinates of two hands are projected to the TransNet's input space (\textit{i.e.}, union of two hand boxes).
Finally, we minimized the L1 distance between the projected and GT 2D coordinates.
Please note that we make the shape parameter of MANO of left and right hands be the same during the weak supervision to minimize the scale ambiguity of each hand.

\section{Architecture of DetectNet}

DetectNet detects left and right hands from an input image $\mathbf{I}_\text{det} \in \mathbb{R}^{3 \times H_\text{det} \times W_\text{det}}$, downsampled from a high-resolution image $\mathbf{I} \in \mathbb{R}^{3 \times 2H_\text{det} \times 2W_\text{det}}$, by predicting two bounding boxes of the left and right hands.
$H_\text{det}=256$ and $W_\text{det}=192$ denote height and width of $\mathbf{I}_\text{det}$, respectively.
The downsampling is necessary to save computational costs.
To this end, we extract the image feature from $\mathbf{I}_\text{det}$ using ResNet-50 and pass the feature to three consecutive deconvolutional layers, which upsample the feature map by 8 times.
We denote the upsampled feature map by $\mathbf{F}_\text{det} \in \mathbb{R}^{C_\text{det} \times H_\text{det}/4 \times W_\text{det}/4}$.
$C_\text{det}=256$ denotes the number of channel of $\mathbf{F}_\text{det}$.
We use the original ResNet-50 after dropping global average pooling (GAP) and following fully-connected layers.
Then, a 1-by-1 convolutional layer takes $\mathbf{F}_\text{det}$ and predicts a 2D heatmap of two hand bounding box centers.
Soft-argmax~\cite{sun2018integral} extracts 2D hand bounding box center coordinates from the 2D heatmap in a differentiable way.
Then, we extract bounding box center features of left and right hands by performing a bilinear interpolation at the box center positions of $\mathbf{F}_\text{det}$.
The extracted bounding box center features of each hand are passed to two fully-connected layers, which produce a scale of the bounding box.
By decoding the bounding box centers and scales, we obtain two bounding boxes of left and right hands.
\section{Architecture of SHNet}

SHNet processes right and left hand images in the same way except for the input and output of the left hand image are flipped to the right hand and flipped back to the left hand, respectively.
Hence, we omit right and left hand notations in the following description. 

SHNet predicts 2.5D joint coordinates $\mathbf{P} \in \mathbb{R}^{J \times 3}$, MANO parameters, and 3D global translation $\mathbf{g} \in \mathbb{R}^3$ from a single hand image $\mathbf{I}_\text{hand} \in \mathbb{R}^{3 \times H_\text{hand} \times W_\text{hand}}$.
$H_\text{hand}=256$ and $W_\text{hand}=256$ denote height and width of $\mathbf{I}_\text{hand}$, respectively.
$J=21$ denotes the number of single hand joints.
MANO parameters include 3D joint rotations $\theta \in \mathbb{R}^{16 \times 3}$ and hand shape parameter $\beta \in \mathbb{R}^{10}$.
The 3D global translation $\mathbf{g}$ is used in two cases: 1) loss calculations and 2) mesh rendering to visualize results.

\noindent\textbf{2.5D joint coordinate estimation.}
ResNet-50 extracts an image feature $\mathbf{F}_\text{hand} \in \mathbb{R}^{C_\text{hand} \times H_\text{hand}/32 \times W_\text{hand}/32}$ from a single hand image $\mathbf{I}_\text{hand}$.
$C_\text{hand}=2048$ denotes the number of channel of $\mathbf{F}_\text{hand}$.
Then, the extracted image feature $\mathbf{F}_\text{hand}$ is passed to a 1-by-1 convolutional layer, which outputs $JD$-dimensional feature map, where $D=8$ denotes discretized depth size.
The feature map is reshaped to the dimension of $\mathbb{R}^{J \times D \times H_\text{hand}/32 \times W_\text{hand}/32}$, which is a 3D heatmap of hand joints.
Soft-argmax~\cite{sun2018integral} extracts 2.5D joint coordinates $\mathbf{P}$ from the 3D heatmap.

\noindent\textbf{MANO parameter regression.}
SHNet firsts reduces the channel dimension of $\mathbf{F}_\text{hand}$ from 2048 to 512 to reduce computational costs.
Then, SHNet extracts joint features by performing a bilinear interpolation at the $(x,y)$ position of the 512-dimensional feature map.
The joint features contain essential articulation information about hand joints.
Finally, a single linear layer outputs MANO pose parameter $\theta$ from a concatenation of the joint features with 2.5D joint coordinates.
The pose parameter $\theta$ is initially estimated in the 6D rotational representation~\cite{zhou2019continuity} and transformed to the axis-angle representation.
The MANO shape parameter $\beta$ and 3D global translation $\mathbf{g}$ are estimated from GAPed $\mathbf{F}_\text{hand}$.

\section{Implementation details} 

PyTorch~\cite{paszke2017automatic} is used for implementation. 
For the training, we use Adam optimizer~\cite{kingma2014adam} with a mini-batch size of 64.
Data augmentations, including scaling, rotation, random horizontal flip, and color jittering, are performed during the training.
The initial learning rate is set to $10^{-4}$ and reduced by a factor of 10 at the \nth{4} epoch.
All other details will be available in our codes.
\section{Limitations}

\begin{figure}[t]
\begin{center}
\includegraphics[width=\linewidth]{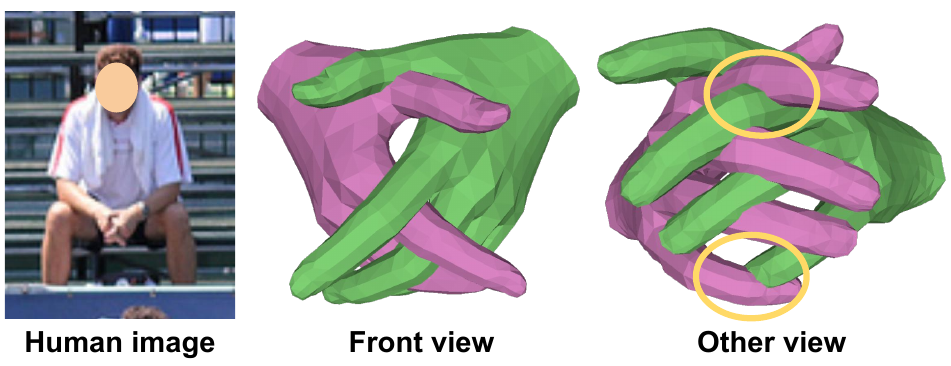}
\end{center}
\vspace*{-5mm}
\caption{
Failure case of our InterWild.
}
\label{fig:limitation}
\end{figure}

Figure~\ref{fig:limitation} shows failure case of ours, highlighted by yellow circles.
It recovers the 3D mesh of the front view correctly; however, there are collisions in the 3D meshes of other views.
This can be addressed by introducing a collision avoidance loss function, similar to IHMR~\cite{rong2021monocular}.
We leave this as our one of future works.

\clearpage

{\small
\bibliographystyle{ieee_fullname}
\bibliography{bib}
}

\end{document}